\documentclass[10pt,twocolumn,letterpaper]{article}

\usepackage{cvpr}
\usepackage{times}
\usepackage{epsfig}
\usepackage{graphicx}
\usepackage{amsmath}
\usepackage{amssymb}

\usepackage{subfigure}
\usepackage{mathrsfs}
\usepackage{amsthm}
\usepackage{algorithm}
\usepackage{algorithmic}
\usepackage{enumerate}
\usepackage{multirow}

\usepackage{enumitem}
\usepackage{makecell}
\usepackage{tabulary}
\usepackage{pifont}

\usepackage[breaklinks=true,bookmarks=false]{hyperref}

\cvprfinalcopy % *** Uncomment this line for the final submission

\def\cvprPaperID{3628} % *** Enter the CVPR Paper ID here

\ifcvprfinal\fi
\begin{document}

%%%%%%%%% TITLE
\title{Transferrable Prototypical Networks for Unsupervised Domain Adaptation}

\author{Yingwei Pan $^{\dag}$, Ting Yao $^{\dag}$, Yehao Li $^{\ddag}$, Yu Wang $^{\dag}$, Chong-Wah Ngo $^{\S}$, and Tao Mei $^{\dag}$ \\
{\small\centering$^{\dag}$ JD AI Research, Beijing, China}\\
{\small\centering$^{\ddag}$ Sun Yat-sen University, Guangzhou, China}\\
{\small\centering$^{\S}$ City University of Hong Kong, Kowloon, Hong Kong}\\
{\tt\scriptsize \{panyw.ustc, tingyao.ustc, yehaoli.sysu, feather1014\}@gmail.com, cscwngo@cityu.edu.hk, tmei@live.com}
}

\input{def.set}
\maketitle
\thispagestyle{empty}

%%%%%%%%% ABSTRACT
\begin{abstract}
  In this paper, we introduce a new idea for unsupervised domain adaptation via a remold of Prototypical Networks, which learn an embedding space and perform classification via a remold of the distances to the prototype of each class. Specifically, we present Transferrable Prototypical Networks (TPN) for adaptation such that the prototypes for each class in source and target domains are close in the embedding space and the score distributions predicted by prototypes separately on source and target data are similar. Technically, TPN initially matches each target example to the nearest prototype in the source domain and assigns an example a ``pseudo" label. The prototype of each class could then be computed on source-only, target-only and source-target data, respectively. The optimization of TPN is end-to-end trained by jointly minimizing the distance across the prototypes on three types of data and KL-divergence of score distributions output by each pair of the prototypes. Extensive experiments are conducted on the transfers across MNIST, USPS and SVHN datasets, and superior results are reported when comparing to state-of-the-art approaches. More remarkably, we obtain an accuracy of 80.4\% of single model on VisDA 2017 dataset.
\end{abstract}

%%%%%%%%% BODY TEXT
\section{Introduction}
The recent advances in deep neural networks have convincingly demonstrated high capability in learning vision models on large datasets. For instance, an ensemble of residual nets \cite{he2016resnet} achieves 3.57\% top-5 error on the ImageNet test set, which is even lower than 5.1\% of the reported human-level performance. The achievements rely heavily on the requirement to have large quantities of annotated data for deep model learning. However, performing intensive manual labeling on a new dataset is expensive and time-consuming. A valid question is why not recycling off-the-shelf learnt knowledge/models in source domain for new domain(s). The difficulty originates from the domain gap \cite{yao2012predicting} that may adversely affect the performance especially when the source and target data distributions are very different. An appealing way to address this challenge would be unsupervised domain adaptation, which aims to utilize labeled examples or learnt models in the source domain and the large number of unlabeled examples in the target domain to generalize a target model.

A common practice in unsupervised domain adaptation is to align data distributions between source and target domains or build invariance across domains by minimizing domain shift through measures such as correlation distances \cite{sun2016fru,yao2015semi} or maximum mean discrepancy \cite{tzeng2014deep}. In this paper, we explore general-purpose and task-specific domain adaptations under the framework of Prototypical Networks \cite{snell2017prototypical}. The design of prototypical networks assumes the existence of an embedding space in which the projections of samples in each class cluster around a single prototype (or centroid). The classification is then performed by computing the distances to prototype representations of each class in the embedding space. In this way, the general-purpose adaptation is to represent each class distribution by a prototype and match the prototypes of each class in the embedding space learnt on the data from different domains. The inspiration of task-specific adaptation is from the rationale that the target data should be classified correctly by the task-specific model when the source and target distributions are well aligned. In the context of prototypical networks, task-specific adaptation is equivalent to adapting the score distributions produced by prototypes in different domains.

By consolidating the idea of general-purpose adaptation and task-specific adaptation into unsupervised domain adaptation, we present a novel Transferrable Prototypical Networks (TPN) architecture. Ideally, TPN is to learn a non-linear mapping (a neural network) of the input examples into an embedding space, in which the representations are invariant across domains. Specifically, TPN takes a batch of labeled source and unlabeled target examples, compares each target example to each of the prototypes computed on source data, and assigns the label of the nearest prototype as a ``pseudo" label to each target example. As such, the general-purpose adaptation is then formulated to minimize the distances between the prototypes measured on source data, target data with pseudo labels, and source plus target data. That is to alleviate domain discrepancy on class level. In task-specific adaptation, we utilize a softmax over distances of the embedding of each example to the prototypes as the classifier. The KL-divergence is exploited to model the mismatch of score distribution by classifiers on prototypes computed in each domain or their combination. In this case, domain discrepancy is amended on sample level. The whole TPN is end-to-end trained by minimizing the classification loss on labeled source data plus the two adaptation terms, and switching the learning from batch to batch. At inference stage, each prototype is computed as a priori. A test target example is projected into the embedding space to compare to each prototype and the outputs of softmax are taken as predictions.

\section{Related Work}
Inspired by the recent advances in image representation using deep convolutional neural networks (DCNNs), a few deep architecture based methods have been proposed for unsupervised domain adaptation. In particular, one common deep solution for unsupervised domain adaptation is to guide the feature learning in DCNNs by minimizing the domain discrepancy with Maximum Mean Discrepancy (MMD) \cite{gretton2012kernel}. MMD is an effective non-parametric metric for the comparisons between the distributions of source and target domains. \cite{tzeng2014deep} is one of early works that incorporates MMD into DCNNs with regular supervised classification loss on source domain to learn both semantically meaningful and domain invariant representation. Later in \cite{long2015learning}, Long \emph{et al.} simultaneously exploit transferability of features from multiple layers via the multiple kernel variant of MMD. The work is further extended by adapting classifiers through a residual transfer module in \cite{long2016unsupervised}. Most recently, \cite{long2017deep} explores domain shift reduction in joint distributions of the network activation of multiple task-specific layers.

Another branch of unsupervised domain adaptation in DCNNs is to exploit the domain confusion by learning a domain discriminator \cite{ganin2015unsupervised,long2018deep,tzeng2015simultaneous,tzeng2017adversarial,zhang2018fully}. Here the domain discriminator is designed to predict the domain (source/target) of each input sample and is trained in an adversarial fashion, similar to GANs \cite{goodfellow2014generative}, for learning domain invariant representation. For example, \cite{tzeng2015simultaneous} devises a domain confusion loss measured in domain discriminator for enforcing the learnt representation to be domain invariant. Similar in spirit, Ganin \emph{et al.} explore such domain confusion problem as a binary classification task and optimize the domain discriminator via a gradient reversal algorithm in \cite{ganin2015unsupervised}. Coupled GANs \cite{liu2016coupled} directly applies GANs into domain adaptation problem to explicitly reduce the domain shifts by learning a joint distribution of multi-domain images. Recently, \cite{tzeng2017adversarial} combines adversarial learning with discriminative feature learning for unsupervised domain adaptation. Most recently, \cite{volpi2018adversarial} extends domain discriminator by learning domain-invariant feature extractor and performing feature augmentation.

In summary, our approach belongs to domain discrepancy based methods. Similar to previous approaches \cite{long2017deep,tzeng2014deep}, our TPN leverages additional unlabeled target data for learning task-specific classifiers. The novelty is on the exploitation of multi-granular domain discrepancy in Prototypical Networks, at class-level and sample-level, that has not been fully explored in the literature. Class-level domain discrepancy is reduced by learning similar prototypes of each class in different domains, while sample-level discrepancy is by enforcing similar score distributions across prototypes of different domains.

\section{Unsupervised Domain Adaptation}
Our Transferrable Prototypical Networks (TPN) is to remould Prototypical Networks towards the scenario of unsupervised domain adaptation by jointly bridging the domain gap via minimizing multi-granular domain discrepancies, and constructing classifiers with unlabeled target data and labeled source data. The classifiers in Prototypical Networks are typically achieved by measuring distances between the example and prototype of each class, which can be flexibly adapted across domains by only updating prototypes in a specific domain. To learn transferrable representations in Prototypical Networks, TPN firstly utilizes the classifiers learnt on source-only data to directly predict the pseudo labels of unlabeled target data and thus produces another two kinds of prototype-based classifiers constructed in target-only and source-target data. The training of TPN is then performed simultaneously by classifying each source sample as correct class and reducing multi-granular domain discrepancy at class level \& sample level. The class-level domain discrepancy is reduced via matching the prototypes of each class, and the sample-level domain discrepancy is minimized by enforcing the score distributions over classes of each sample synchronized, across different domains. We alternate the above two steps in each training iteration and optimize the whole TPN in an end-to-end fashion.

\begin{figure*}[!tb]
    \centering {\includegraphics[width=0.75\textwidth]{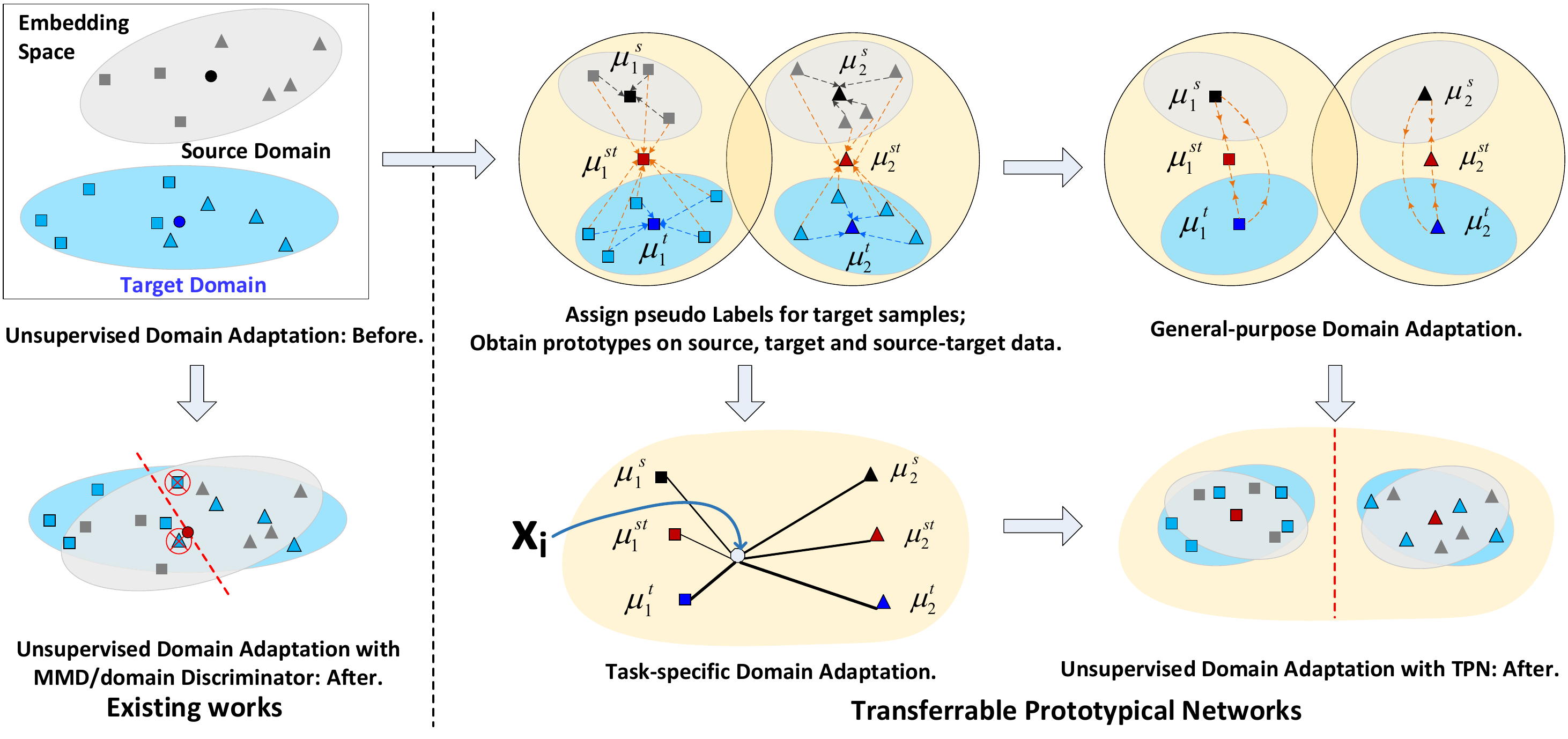}}
    \vspace{-0.05in}
    \caption{The intuition behind existing unsupervised domain adaptation models with MMD \cite{long2015learning} or domain discriminator \cite{tzeng2015simultaneous} and our Transferrable Prototypical Networks (TPN) (better viewed in color). Most of the existing models aim to reduce the domain shift by measuring the holistic domain discrepancy/domain confusion over source and target data, while leaving the domain discrepancy of each class or the relations between samples and classifiers unexploited. In contrast, our TPN tackles this problem from the viewpoint of both general-purpose and task-specific adaptation to measure the multi-granular domain discrepancy at class level and sample level, respectively. In particular, TPN initially matches each unlabeled target sample to the nearest prototype in the source domain and assigns each target example a ``pseudo" label. Next, the prototype of each class is computed on source-only, target-only and source-target data. The general-purpose adaptation is then performed to push the prototype of each class computed in each domain to be close in the embedding space. Meanwhile, we perform the task-specific adaptation to align the score distributions produced by prototypes obtained in different domains for each sample. The whole TPN is trained by minimizing the supervised classification loss on labeled source data plus the general-purpose and task-specific adaptation terms in an end-to-end manner.}
    \label{fig:1}
    \vspace{-0.2in}
\end{figure*}

\subsection{Preliminary---Prototypical Networks}
Prototypical Networks is preliminarily proposed in \cite{snell2017prototypical} to construct an embedding space in which points cluster around a single prototype representation of each class. In particular, given a set with $N$ labeled samples $\mathcal{S} = \{(x_i,y_i)\}^N_{i=1}$ belonging to $C$ categories, where ${y_i} \in \left\{ {1,2,...,C} \right\}$ is the class label of sample $x_i$. The objective is to learn an embedding function $f\left( {x_i;\theta } \right):x_i \to {\mathbb{R}}^m$ for transforming each input sample into a $m$-dimensional embedding space through a deep architecture of Prototypical Networks, where $\theta$ represents the learnable parameters. To convey the high-level description of the class as meta-data, the prototype of each class is defined by taking the average of all embedded samples belonging to that class:
\begin{equation}\label{Eq:Eq1}\small
{{\mu}_c}  = \frac{1}{{\left| {{\mathcal{S}_c}} \right|}}\sum\limits_{{x_i} \in {\mathcal{S}_c}} {f\left( {{x_i};\theta } \right)},
\end{equation}
where ${\mathcal{S}_c}$ denotes the set of samples from class $c$. Given a query sample $x_i$, Prototypical Networks directly produce its score distribution ${\bf{P}}_i \in {\mathbb{R}}^C$ over $C$ classes via a softmax function on distances to the prototypes, whose $c$-th element is the probability of $x_i$ belonging to class $c$:
\begin{equation}\label{Eq:Eq2}\small
{{\bf{P}}_{ic}} = p\left( {y_i = c|x_i} \right) = \frac{{{e^{ - d\left( {f\left( {{x_i};\theta } \right),{\mu _c}} \right)}}}}{{\sum\nolimits_{c'} {{e^{ - d\left( {f\left( {{x_i};\theta } \right),{\mu _{c'}}} \right)}}} }},
\end{equation}
where $d\left( \cdot \right)$ is the distance function (e.g., Euclidean distance as in \cite{snell2017prototypical}) between query sample and the prototype. The training of Prototypical Networks is performed by minimizing the negative log-likelihood probability of assigning correct class label $c$ to this sample:
\begin{equation}\label{Eq:Eq3}\small
{L_S}\left( x_i \right) =  - \log p\left( {y_i = c|x_i} \right).
\end{equation}

\subsection{Problem Formulation}
In unsupervised domain adaptation, we are given $N_s$ labeled samples $\mathcal{S}^s = \{(x^s_i,y^s_i)\}^{N_s}_{i=1}$ in the source domain and $N_t$ unlabeled samples $\mathcal{S}^t = \{x^t_i\}^{N_t}_{i=1}$ in the target domain. Based on the widely adopted assumption of the existence of a shared feature space for source and target domains in \cite{long2017deep,pan2008transfer,tzeng2015simultaneous}, the ultimate goal of this task is to design an embedding function $f\left( {x_i;\theta } \right)$ which formally reduces domain shifts in the shared feature space and enables learning of both transferrable representations and classifiers depending on $\mathcal{S}^s$ and $\mathcal{S}^t$. Different from the existing transfer techniques \cite{long2017deep,long2016unsupervised} which are typically composed of two cascaded networks for learning domain-invariant features and target-discriminative classifiers respectively, we consider unsupervised domain adaptation in the framework of Prototypical Networks. Such framework naturally unifies the learning of features and classifiers into one network by constructing classifiers purely on the prototype of each class. This design reflects a very simple inductive bias that is beneficial in domain adaptation regime. Specifically, to make Prototypical Networks transferrable across domains, two adaptation mechanisms are devised to align distributions of source and target domains through reducing multi-granular (i.e., class-level and sample-level) domain discrepancies. In between, the general-purpose adaptation matches the prototypes of each class and the task-specific adaptation enforces similar score distributions over classes of each sample, across different domains, as shown in Figure \ref{fig:1}.

\subsection{General-purpose Domain Adaptation}
Most existing works resolve unsupervised domain adaptation by minimizing the domain discrepancy between source and target data distributions with MMD \cite{tzeng2014deep}, or maximizing the domain confusion across domains via a domain discriminator \cite{tzeng2015simultaneous}. Both of the domain discrepancy and domain confusion terms are measured over the entire source and target data, irrespective of the specific class of each sample. Moreover, the domain discrepancy has been seldom exploited across domains for each class, possibly because measuring such class-level domain discrepancy needs the labels of both source and target samples, while in typical unsupervised domain adaptation settings, no label is provided for target samples.

Inspired from self-labeling \cite{lee2013pseudo,saito2017asymmetric} for domain adaptation, we directly utilize prototype-based classifier learnt on labeled source data for matching each target sample to the nearest prototype in the source domain, and then assign the target sample a ``pseudo" label. As such, all the target samples $\mathcal{\hat S}^t = \{(x^t_i,{\hat y}^t_i)\}^{N_t}_{i=1}$ are with pseudo labels. After obtaining the real/pseudo labels of source/target data, three kinds of classifiers (i.e., prototypes $\left\{ {\mu _{_c}^s} \right\}$, $\left\{ {\mu _{_c}^t} \right\}$ and $\left\{ {\mu _{_c}^{st}} \right\}$) could be calculated on source-only data ($\mathcal{S}^s$), target-only data ($\mathcal{\hat S}^t$) and source-target data ($\mathcal{S}^s \cup \mathcal{\hat S}^t$), respectively:
\begin{equation}\label{Eq:Eq4}\small
\begin{array}{l}
\mu _{_c}^s = \frac{1}{{\left| {\mathcal{S}_c^s} \right|}}\sum\limits_{x_i^s \in \mathcal{S}_c^s} {f\left( {x_i^s;\theta } \right)} ,\mu _{_c}^t = \frac{1}{{\left| {\mathcal{\hat S}_c^t} \right|}}\sum\limits_{x_i^t \in \mathcal{\hat S}_c^t} {f\left( {x_i^t;\theta } \right)},\\
\mu _{_c}^{st} = \frac{1}{{\left| {\mathcal{S}_c^s} \right| + \left| {\mathcal{\hat S}_c^t} \right|}}\left( {\sum\limits_{{x^s_i} \in \mathcal{S}_c^s} {f\left( {x_i^s;\theta } \right)}  + \sum\limits_{x_i^t \in \mathcal{\hat S}_c^t} {f\left( {x_i^t;\theta } \right)} } \right),
\end{array}
\end{equation}
where ${\mathcal{S}_c^s}$ and ${\mathcal{\hat S}_c^t}$ denote the sets of source/target samples from the same class $c$.

To measure the class-level domain discrepancy across domains, we take the inspiration from MMD-based transfer techniques \cite{long2017deep,long2016unsupervised} and compute pairwise reproducing kernel Hilbert space (RKHS) distance between the prototypes of the same class from different domains. The basic idea is that if the data distributions of source and target domains are identical, the prototypes of the same class achieved on different domains are the same. Formally, we define the following class-level discrepancy loss as
\begin{equation}\label{Eq:Eq5}\small
\begin{array}{l}
{L_G}\left( {\left\{ {\mu _{_c}^s} \right\},\left\{ {\mu_{_c}^t} \right\},\left\{ {\mu _{_c}^{st}} \right\}} \right) {\buildrel \Delta \over =} \frac{1}{C}\sum\limits_{c = 1}^C {\left\| {\widetilde\mu _{_c}^s - \widetilde\mu _{_c}^t} \right\|_{\mathcal H}^2}\\
~~~~~~~~+ \frac{1}{C}\sum\limits_{c = 1}^C {\left\| {\widetilde\mu _{_c}^s - \widetilde\mu _{_c}^{st}} \right\|_{\mathcal H}^2} + \frac{1}{C}\sum\limits_{c = 1}^C {\left\| {\widetilde\mu _{_c}^t - \widetilde\mu _{_c}^{st}} \right\|_{\mathcal H}^2},
\end{array}
\end{equation}
where $\left\{ {\widetilde\mu _{_c}^s} \right\}$, $\left\{ {\widetilde\mu _{_c}^t} \right\}$ and $\left\{ {\widetilde\mu _{_c}^{st}} \right\}$ denote the corresponding prototypes in reproducing kernel Hilbert space ${\mathcal H}$. By minimizing this term, the prototype of each class computed in each domain will be enforced to be in close proximity in the embedding space, leading to invariant representation distribution across domains in general.

\textbf{Connections with MMD.} MMD \cite{gretton2012kernel} is a kernel two-sample test which measures the distribution difference between source and target data by mapping them into a reproducing kernel Hilbert space. The empirical estimation of MMD is computed by
\begin{equation}\label{Eq:Eq6}\small
\begin{array}{l}
{\mu ^s} = \frac{1}{{\left| {{S^s}} \right|}}\sum\limits_{x_i^s \in {S^s}} {\phi \left( {x_i^s} \right)} ,{\mu ^t} = \frac{1}{{\left| {{S^t}} \right|}}\sum\limits_{x_i^t \in {S^t}} {\phi \left( {x_i^t} \right)} ,\\
{L_{MMD}} = \left\| {{\mu ^s} - {\mu ^t}} \right\|_{\mathcal H}^2,
\end{array}
\end{equation}
where $\phi \left( \cdot \right)$ is the mapping to RKHS ${\mathcal H}$. Taking a close look on the objective of MMD and our class-level discrepancy loss in Eq.(\ref{Eq:Eq5}), we can observe some interesting connections. Concretely, the means of source and target data (i.e., ${\mu ^s}$ and ${\mu ^t}$) measured in MMD can be interpreted as the \emph{holistic} prototype of each domain in RKHS. MMD is then expressed as the RKHS distance between the \emph{holistic} prototypes across domains. Our class-level domain discrepancy, different from MMD, is computed as the RKHS distance across the prototypes of each class from different domains. In other words, a fine-grained alignment of source and target data distributions is performed on class level, instead of simply minimizing the distance between \emph{holistic} prototypes across domains.

\subsection{Task-specific Domain Adaptation}
The general-purpose domain adaptation only enforces similarity in feature distributions, while leaving the relations between samples and task-specific classifiers (i.e., prototypes) unexploited. Furthermore, we devise a new adaptation mechanism, i.e., task-specific adaptation, to reduce sample-level domain discrepancy by aligning the score distributions of different classifiers (i.e., prototypes) across domains for each sample. The rationale of sample-level domain discrepancy is that each source/target sample should be classified correctly by the task-specific classifiers when source and target distributions are well aligned, leading to consistent decisions of classifiers across domains.

In particular, given each source/target sample $x_i$, three score distributions (${{\bf{P}}_i^s}$, ${{\bf{P}}_i^t}$ and ${{\bf{P}}_i^{st}}$) are obtained via three kinds of classifiers (i.e., prototypes $\left\{ {\mu _{_c}^s} \right\}$, $\left\{ {\mu _{_c}^t} \right\}$ and $\left\{ {\mu _{_c}^{st}} \right\}$) learnt on source-only, target-only and source-target data, respectively. To measure the sample-level domain discrepancy, we utilize KL-divergence to evaluate the pairwise distance between the score distributions from different domains. The sample-level discrepancy loss over the source and target samples are defined as
\begin{equation}\label{Eq:Eq7}\scriptsize
\begin{array}{l}
{L_T}\left( {\left\{ {{\bf{P}}_i^s} \right\},\left\{ {{\bf{P}}_i^t} \right\},\left\{ {{\bf{P}}_i^{st}} \right\}} \right) {\buildrel \Delta \over =} \frac{1}{{{\left| {{\mathcal{S}^s}} \right| + \left| {{\mathcal{\hat S}^t}} \right|}}}\sum\limits_{{x_i}} {{D_{KL}}\left( {{\bf{P}}_i^s,{\bf{P}}_i^t} \right)} \\
~~~~~~~~~~~~~~~~~~~~~~~~~~~~~~~~~~~~~~~~~~~~~~~~~+ \frac{1}{{{\left| {{\mathcal{S}^s}} \right| + \left| {{\mathcal{\hat S}^t}} \right|}}}\sum\limits_{{x_i}} {{D_{KL}}\left( {{\bf{P}}_i^s,{\bf{P}}_i^{st}} \right)} \\
~~~~~~~~~~~~~~~~~~~~~~~~~~~~~~~~~~~~~~~~~~~~~~~~~+ \frac{1}{{{\left| {{\mathcal{S}^s}} \right| + \left| {{\mathcal{\hat S}^t}} \right|}}}\sum\limits_{{x_i}} {{D_{KL}}\left( {{\bf{P}}_i^t,{\bf{P}}_i^{st}} \right)} ,\\
{D_{KL}}\left( {{\bf{P}}_i^s,{\bf{P}}_i^t} \right) = \frac{1}{2}\left( {{d_{KL}}\left( {{\bf{P}}_i^s||{\bf{P}}_i^t} \right) + {d_{KL}}\left( {{\bf{P}}_i^t||{\bf{P}}_i^s} \right)} \right),\\
{d_{KL}}\left( {{\bf{P}}_i^s||{\bf{P}}_i^t} \right) = \sum\limits_{c = 1}^C {{\bf{P}}_{ic}^s\log \left( {\frac{{{\bf{P}}_{ic}^s}}{{{\bf{P}}_{ic}^t}}} \right)},
\end{array}
\end{equation}
where ${d_{KL}} \left( \cdot \right)$ is the KL-divergence factor and ${D_{KL}} \left( \cdot \right)$ is the symmetric pairwise KL-divergence.

Please note that different from general-purpose domain adaptation which independently matches the prototypes of each class across different domains, task-specific adaptation simultaneously adapts the prototypes of all classes, pursuing similar score distributions over classes of each sample.

\subsection{Optimization}
The overall training objective of our TPN integrates the supervised classification loss in Eq.(\ref{Eq:Eq3}) and multi-granular discrepancy losses (i.e., class-level discrepancy loss in Eq.(\ref{Eq:Eq5}) and sample-level discrepancy loss in Eq.(\ref{Eq:Eq7})). Hence we obtain the following optimization problem:
\begin{equation}\label{Eq:Eq8}\small
\begin{array}{l}
\mathop {\min }\limits_\theta  \frac{1}{{\left| {{\mathcal{S}^s}} \right|}}\sum\limits_{x_i^s \in {\mathcal{S}^s}} {{L_S}\left( {x_i^s} \right)}  + \alpha {L_G}\left( {\left\{ {\mu _{_c}^s} \right\},\left\{ {\mu _{_c}^t} \right\},\left\{ {\mu _{_c}^{st}} \right\}} \right) \\
~~~~~~~~~~~~~~~~~~~~~~~~+ \beta {L_T}\left( {\left\{ {{\bf{P}}_i^s} \right\},\left\{ {{\bf{P}}_i^t} \right\},\left\{ {{\bf{P}}_i^{st}} \right\}} \right),
\end{array}
\end{equation}
where $\alpha$ and $\beta$ are tradeoff parameters. With this overall loss objective, the crucial goal of the optimization is to learn the deep embedding function $f\left( {x_i;\theta } \right)$, in which the output representations are invariant across domains.

\textbf{Training Procedure.} To address the optimization problem in Eq.(\ref{Eq:Eq8}), we split the training process into two steps: 1) calculate classifier (i.e., prototypes $\left\{ {\mu _{_c}^s} \right\}$) on source domain and perform it to assign pseudo labels to target samples; 2) calculate classifiers (i.e., prototypes $\left\{ {\mu _{_c}^t} \right\}$ and $\left\{ {\mu _{_c}^{st}} \right\}$) on target-only and source-target data, and update $\theta$ with respect to the gradient descent of overall objective function. We alternate the two steps in each training iteration and stop the procedure until a convergence criterion is met. Note that to remedy the error of self-labeling, we only assign pseudo labels to the target examples whose maximized scores are over 0.6 and resample the target examples for labeling in each training iteration to avoid overfitting of pseudo labels. Furthermore, the training process of our TPN is also resistant to the noise of pseudo labels since we iteratively utilize both labeled source examples and pseudo-labeled target examples for learning the embedding function. This procedure not only ensures the accuracy in source domain, but also effectively minimizes class-level and sample-level discrepancy. Such cycle will gradually improve the accuracy in target domain.

\textbf{Inference.} After training TPN, we can obtain the deep embedding function $f\left( {x_i;\theta } \right)$. With this, all the three sets of prototypes ($\left\{ {\mu _{_c}^s} \right\}$, $\left\{ {\mu _{_c}^t} \right\}$ and $\left\{ {\mu _{_c}^{st}} \right\}$) are calculated over the whole training set in advance and stored in memory. Any one of the three prototype sets can be utilized as the final classifier for classifying test target sample at the testing stage. We empirically verified that the performance is not sensitive to the selection of prototypes\footnote{\small The accuracy constantly fluctuates within 0.002 when using different set of prototypes for four domain shifts in experiments.}, which implicitly reveals the domain invariant characteristic of the learnt feature representation. Hence, given a test target sample, we compute its embedding representation via $f\left( {x_i;\theta } \right)$ and compare the distances to prototypes of each class to output the final prediction scores.

\subsection{Theoretical Analysis}
We formalize the error bound of TPN by an extension of the theory in \cite{ben2010theory}. As TPN performs training on a mixture of labeled source examples and target samples with pseudo labels, the classification error is naturally considered as the linear weighted sum of errors in source and target domain. Denote $y^s$ and $\hat y^t$ as the ground truth labels of source examples and the pseudo labels of target samples, respectively, and $h$ as a hypothesis. The error is then formally written as
\begin{equation}\small
\epsilon_{\gamma}(h)=\gamma \epsilon_{t}(h,\hat y^t)+(1-\gamma)\epsilon_{s}(h,y^s)~,
\label{ErrorTPN}
\end{equation}
where $\gamma$ is the tradeoff parameter. The term $\epsilon_{t}(h,\hat y^t)=E_{x\sim \mathcal{D}^t}[|h(x)-\hat y^t|]$ and $\epsilon_{s}(h,y^s)=E_{x\sim \mathcal{D}^s}[|h(x)-y^s|]$ represents the expected error over the sample distribution of target domain $\mathcal{D}^t$ and source domain $\mathcal{D}^s$ with respect to pseudo labels and ground truth labels, respectively.

Next, a valid question is how close the error $\epsilon_{\gamma}(h)$ is to an oracle error $\epsilon_{t}(h, y^t)$ that evaluates the classifier learnt on the ground truth labels $y^t$ of the target examples. The closer the two losses are, the more desirable the domain adaptation performs. The following Lemma proves that the difference between the two losses could be bounded for our TPN.
\begin{lemma}\label{lemmaTPN}
Let $h$ be a hypothesis in class $\mathcal{H}$. Then
\begin{equation}\scriptsize
\left|\epsilon_{\gamma}(h)-\epsilon_{t}(h,y^t)\right| \le(1-\gamma) (\frac{1}{2}d_{\mathcal{H}\Delta{H}}(\mathcal{D}^s,\mathcal{D}^t)+\lambda)+\gamma\rho,
\label{TPNbounderror}
\end{equation}
where $d_{\mathcal{H}\Delta{H}}(\mathcal{D}^s,\mathcal{D}^t)= 2 \sup_{h,h' \in\mathcal{H}} |\epsilon_t(h,h')-\epsilon_s(h,h')|$ measures the domain discrepancy in the hypothesis space $\mathcal{H}$. $\rho$ denotes the ratio of target examples with false pseudo labels. $\lambda=\epsilon_s(h^*,y^s)+\epsilon_{t}(h^*,y^{ t})$ is the combined error in two domains of the joint ideal hypothesis $h^*$, which is the optimal hypothesis by minimizing the combined error:
\begin{equation}\small
h^*=\argmin\epsilon_s(h,y^s)+\epsilon_t(h,y^t).
\end{equation}
\end{lemma}
Lemma \ref{lemmaTPN} decomposes the bound into three terms: domain discrepancy $d_{\mathcal{H}\Delta{H}}(\mathcal{D}^s,\mathcal{D}^t)$ measured by the disagreement of hypothesis in the space $\mathcal{H}$, the error $\lambda$ of the ideal joint hypothesis and the ratio $\rho$ of the noise in pseudo labels. In TPN, the first term is assessed through quantifying class-level discrepancy of prototypes and sample-level discrepancy over score distributions across different domains. As stated in \cite{ben2010theory}, when the combined error $\lambda$ of the joint ideal hypothesis is large, there is no classifier that performs well on both domains. Instead, in the most relevant cases for domain adaptation, $\lambda$ is usually considered to be negligibly small and thus the second term can be disregarded. Furthermore, in each iteration, TPN searches for the optimal hypothesis and improves the accuracy of pseudo-label prediction on target examples. The increase of correct pseudo labels in turn benefits the reduction of domain discrepancy. We will empirically verify that the third term $\rho$ of the noise in pseudo labels is iteratively decreased in Section \ref{ssec:EA}. As such, TPN constantly tightens the bound in Eq.(\ref{TPNbounderror}).

\section{Experiments}
We conduct extensive evaluations of TPN for unsupervised domain adaptation from four domain shifts, including three Digits image transfer across three Digits datasets (i.e., MNIST \cite{lecun1998gradient}, USPS \cite{friedman2001elements} and SVHN \cite{netzer2011reading}) and one synthetic-to-real image transfer on VisDA 2017 dataset \cite{visda2017}.

\subsection{Datasets and Experimental Settings}
\textbf{Datasets.}
The MNIST (M) and USPS (U) image datasets are both handwritten Digits datasets containing 10 classes of digits. The MNIST dataset consists of 70$k$ images and the USPS dataset includes 9.3$k$ images. Unlike the two, the SVHN (S) dataset is a real-world Digits dataset of house numbers in Google street view images and contains 100$k$ cropped Digits images. The VisDA 2017 dataset is the largest synthetic-to-real object classification dataset to date with over 280$k$ images in the training, validation and testing splits (domains). All the three domains share the same 12 object categories. The training domain consists of 152$k$ synthetic images which are generated by rendering 3D models of the same object categories from different angles and under different lighting conditions. The validation domain includes 55$k$ images by cropping object in real images from COCO \cite{lin2014microsoft}. The testing domain contains 72$k$ images cropped from video frames in YT-BB \cite{real2017youtube}.

\textbf{Digits Image Transfer.} Following \cite{tzeng2017adversarial}, we consider three directions: M $\rightarrow$ U, U $\rightarrow$ M and S $\rightarrow$ M, for unsupervised domain adaptation among Digits datasets. For the transfer between MNIST and USPS, we sample 2$k$ images from MNIST and 1.8$k$ images from USPS as in \cite{tzeng2017adversarial}. For S $\rightarrow$ M, the two training sets are fully utilized. In addition, the CNN architecture for the three Digits image transfer tasks is a simple modified version of \cite{lecun1998gradient} (2 conv-layer LeNet), which is also exploited in \cite{tzeng2017adversarial}.

\textbf{Synthetic-to-Real Image Transfer.} The second experiment was conducted over the most challenging synthetic-to-real image transfer task in VisDA 2017. As the annotations of the testing data in VisDA are not publicly available, we take the training data (i.e., synthetic images) as source domain and the validation data (i.e., cropped COCO images) as target domain. Moreover, we adopt 50-layer ResNet \cite{he2016resnet} pre-trained on ImageNet \cite{ILSVRC15} as our basic CNN~structure.

\textbf{Implementation Details.} The two tradeoff parameters $\alpha$ and $\beta$ in Eq.(\ref{Eq:Eq8}) are simply set as 1. A common practice in unsupervised domain adaption is the lack of annotations in target domain, making the parameters unable to be well estimated. As such, we directly fix the tradeoff parameters in all the experiments. We strictly follow \cite{french2018self,tzeng2017adversarial} and set the embedding size $m$ as 10/512 for Digits/synthetic-to-real image transfer. We mainly implement TPN based on Caffe \cite{Jia:MM14}. Specifically, the network weights are trained by ADAM \cite{kingma2014adam} with 0.0005 weight decay and 0.9/0.999 momentum for Digits/synthetic-to-real image transfer. The learning rate and mini-batch size are set as 0.0002/0.00001 and 128/60 for Digits/synthetic-to-real image transfer. The maximum training iteration is set as 70$k$ for all the experiments. Moreover, following \cite{tzeng2017adversarial}, we pre-train TPN on labeled source data. For Digits image transfer tasks, we adopt the classification accuracy on target domain as evaluation metric. For synthetic-to-real image transfer, we measure the per-category classification accuracy on target domain. The final metric is the average of accuracy over all~categories.

\textbf{Compared Methods.} To empirically verify the merit of our TPN, we compare the following approaches: (1) \textbf{Source-only} directly exploits the classification model trained on source domain to classify target samples. (2) \textbf{RevGrad} \cite{ganin2015unsupervised} treats domain confusion as a binary classification task and trains the domain discriminator via gradient reversal. (3) \textbf{DC} \cite{tzeng2015simultaneous} explores a Domain Confusion loss measured in domain discriminator for unsupervised domain adaptation. (4) \textbf{DAN} \cite{long2015learning} utilizes multiple kernel variant of MMD to align feature representations from multiple layers. (5) \textbf{RTN} \cite{long2016unsupervised} extends DAN by adapting classifiers through a residual transfer module. (6) \textbf{ADDA} \cite{tzeng2017adversarial} designs an unified unsupervised domain adaptation model based on adversarial learning objectives. (7) \textbf{JAN} \cite{long2017deep} learns a transfer model by aligning joint distributions of the network activation of multiple layers across domains. (8) \textbf{MCD} \cite{saito2018maximum} aligns distributions of source and target domains by utilizing the task-specific decision boundaries. (9) \textbf{S-En} \cite{french2018self} explores the mean teacher variant of temporal ensembling \cite{tarvainen2017mean} for unsupervised domain adaptation. (10) \textbf{TPN} is the proposal in this paper. Moreover, two slightly different settings of TPN are named as \textbf{TPN$_{gen}$} and \textbf{TPN$_{task}$} which are trained with only general-purpose and task-specific adaptation, respectively. (11) \textbf{Train-on-target} is an oracle run that trains the classifier on all labeled target samples.

\begin{table*}\small
\caption{\small Classification accuracy (\%) of different methods for (a) Digits image transfer across MNIST (M), USPS (U) and SVHN (S), and (b) Synthetic-to-real image transfer on VisDA 2017 dataset. For digits image transfer, $^{*}$ indicates the results are directly drawn from \cite{tzeng2017adversarial}. For synthetic-to-real image transfer, $^{\dagger}$ indicates the results are directly drawn from \cite{saito2018maximum} and \cite{french2018self}, respectively.}
\vspace{-0.00in}
\subtable[Digits image transfer.]{
  \setlength\tabcolsep{0.1pt}
  \begin{tabular}{l | c c c}
  \Xhline{2\arrayrulewidth}
  Method & M $\rightarrow$ U  & U $\rightarrow$ M & S $\rightarrow$ M \\
  \hline \hline
  Source-only$^{*}$ & 75.2 & 57.1 & 60.1 \\\hline
  RevGrad \cite{ganin2015unsupervised}$^{*}$        & 77.1 & 73.0 & 73.9 \\
  DC  \cite{tzeng2015simultaneous}$^{*}$ & 79.1 & 66.5 & 68.1 \\
  DAN \cite{long2015learning}     & 80.3 & 77.8 & 73.5 \\
  RTN \cite{long2016unsupervised} & 82.0 & 81.2 & 75.3 \\
  ADDA \cite{tzeng2017adversarial}$^{*}$ & 89.4 & 90.1 & 76.0 \\
  JAN  \cite{long2017deep}        & 84.4 & 83.4 & 78.4 \\
  MCD  \cite{saito2018maximum}    & 90.0 & 88.5 & 83.3  \\
  \hline
  TPN$_{gen}$  & 91.3 & 93.5 & 90.2 \\
  TPN$_{task}$ & 88.1 & 88.0 & 88.8 \\
  TPN          & \textbf{92.1} & \textbf{94.1} & \textbf{93.0} \\\hline
  Train-on-target & 92.3 & 96.8 & 96.8 \\
  \Xhline{2\arrayrulewidth}
  \end{tabular}
  \label{tab:digitresult}
}
~
\subtable[Synthetic-to-real image transfer.]{
  \setlength\tabcolsep{1.5pt}
  \begin{tabular}{l | c c c c c c c c c c c c | c}
  \Xhline{2\arrayrulewidth}
  Method & plane & bcycl & bus & car & horse & knife & mcycl & person & plant & sktbrd & train & truck & mean \\
  \hline\hline
  Source-only & 70.6 & 51.8 & 55.8 & 68.9 & 77.9 & 7.6  & 93.3 & 34.5 & 81.1 & 27.9 & 88.6 & 5.6  & 55.3 \\\hline
  RevGrad \cite{ganin2015unsupervised}   & 75.9 & 70.5 & 65.3 & 17.3 & 72.8 & 38.6 & 58.0 & 77.2 & 72.5 & 40.4 & 70.4 & 44.7  & 58.6  \\
  DC  \cite{tzeng2015simultaneous} & 63.6 & 38.4 & 71.2 & 61.4 & 71.4 & 10.9 & 86.6 & 43.5 & 70.2 & 47.7 & 79.8 & 21.6 & 55.5 \\
  DAN \cite{long2015learning}      & 61.7 & 54.8 & 77.7 & 32.2 & 75.0 & 80.8 & 78.3 & 46.9 & 66.9 & 34.5 & 79.6 & 29.1 & 59.8 \\
  RTN \cite{long2016unsupervised}  & 79.5 & 59.6 & 78.0 & 47.4 & 82.7 & \textbf{82.0} & 84.7 & 54.7 & 81.6 & 34.5 & 74.2 & 6.6  & 63.8 \\
  JAN  \cite{long2017deep}         & 92.1 & 66.4 & 81.4 & 39.6 & 72.5 & 70.5 & 81.5 & 70.5 & 79.7 & 44.6 & 74.2 & 24.6 & 66.5 \\
  MCD  \cite{saito2018maximum}$^{\dagger}$  & 87.0 & 60.9 & \textbf{83.7} & 64.0 & 88.9 & 79.6 & 84.7 & 76.9 & 88.6 & 40.3 & 83.0 & 25.8 & 71.9 \\
  \hline
  TPN$_{gen}$ & \textbf{94.5} & \textbf{86.8} & 76.8 & 49.7 & 92.1 & 12.5 & 84.7 & 75.2 & 92.1 & \textbf{86.8} & 84.1 & 47.4 & 73.6 \\
  TPN$_{task}$ & 89.2 & 62.8 & 71.7 & \textbf{83.5} & 90.6 & 24.6 & 88.8 & \textbf{91.1} & 89.8 & 74.7 & 69.1 & 36.1 & 72.7 \\
  TPN        & 93.7 & 85.1 & 69.2 & 81.6 & \textbf{93.5} & 61.9 & \textbf{89.3} & 81.4 & \textbf{93.5} & 81.6 & \textbf{84.5} & \textbf{49.9} & \textbf{80.4} \\\hline
  S-En+Mini-aug \cite{french2018self}$^{\dagger}$ &92.9 &84.9 &71.6 &41.2 &88.8 &92.4 &67.5 &63.5 &84.5 &71.8 &83.2 &48.1 & 74.2\\
  S-En+Test-aug \cite{french2018self}$^{\dagger}$ &96.3 &87.9 &84.7 &55.7 &95.9 &95.2 &88.6 &77.4 &93.3 &92.8 &87.5 &38.2 & 82.8\\
  \hline
  Train-on-target & 99.5 & 91.9 & 97.3 & 96.8 & 98.3 & 98.5 & 94.1 & 96.2 & 99.0 & 98.2 & 97.9 & 82.3 & 95.8 \\
  \Xhline{2\arrayrulewidth}
  \end{tabular}
  \label{tab:visdaresult}
}
\vspace{-0.2in}
\end{table*}

\subsection{Performance Comparison}
\textbf{Digits Image Transfer.} Table \ref{tab:digitresult} shows the performance comparisons on three transfer directions among Digits datasets. Overall, the results across three adaptations consistently indicate that our proposed TPN achieves superior performances against other state-of-the-art techniques including MMD based models (DAN, RTN, JAN) and domain discriminator based approaches (RevGrad, DC, ADDA, MCD). In particular, the accuracy of TPN can achieve 92.1\% and 94.1\% on the adaptation of M $\rightarrow$ U and U $\rightarrow$ M, making the absolute improvement over the best competitor ADDA by 2.7\% and 4\%, respectively, which is generally considered as a significant progress on the adaptation between MNIST and USPS. It is noteworthy that compared to JAN, our TPN also promotes the classification accuracy evidently on the harder transfer S $\rightarrow$ M, where the source and target domains are substantially different. The results in general highlight the key importance of exploring both class-level and sample-level domain discrepancy via general-purpose and task-specific adaptation in unsupervised domain adaptation, leading to more domain-invariant feature representations.

The performances of Source-only which trains the classifier only on labeled source data could be regarded as a lower bound without domain adaptation. By additionally incorporating the domain adaptation term (MMD/domain discriminator), RevGrad, DC, DAN, RTN, ADDA, JAN and MCD lead to a large performance boost over Source-only, which basically indicates the advantage of measuring the domain discrepancy/domain confusion over the source and target data. Furthermore, the performances of them on harder transfer S $\rightarrow$ M are much lower than our TPN$_{gen}$ and TPN$_{task}$ which exploits the class-level/sample-level domain discrepancy in Prototypical Networks by matching the prototypes across domains for each class and score distributions of different classifiers (i.e., prototypes) for each sample, respectively. This confirms the effectiveness of leveraging class-level and sample-level domain discrepancy in general-purpose and task-specific adaptation, especially between more distinct domains. For the two easy transfer tasks between MNIST and USPS, TPN$_{task}$ is inferior to ADDA, MCD and TPN$_{gen}$, which indicates that solely matching score distributions of each sample might inject noise more easier than domain discriminator/class-level domain discrepancy on transfer task across similar domains. In addition, by simultaneously utilizing both general-purpose and task-specific adaptation, our TPN consistently boosts up the performances on all the three Digits image transfer tasks. The results demonstrate the advantage of jointly leveraging multi-granular domain discrepancy at class level and sample level for unsupervised domain adaptation. Note that we exclude the published results of S-En in this comparison as S-En is originally built with deeper CNNs (i.e., 9 conv layers) on Digits image datasets and our TPN is based on 2 conv-layer LeNet. When equipped with the same CNNs in S-En, the accuracy of our TPN is boosted up to 98.6\% on M $\rightarrow$ U which is higher than 98.3\% of~S-En.

\textbf{Synthetic-to-Real Image Transfer.}
The performance comparisons for synthetic-to-real image transfer task on VisDA 2017 dataset are summarized in Table \ref{tab:visdaresult}. Here the results of S-En are all reported on the setting with multiple data augmentations (DA). Our TPN performs consistently better than other runs without any DA involved. In particular, the mean accuracy across all the 12 categories can reach 80.4\%, making the absolute improvement over JAN by 13.9\%. Similar to the observations on the hard Digits image transfer S $\rightarrow$ M, TPN$_{gen}$ and TPN$_{task}$ exhibit better performance than JAN by taking class-level and sample-level domain discrepancy into account for unsupervised domain adaptation. In addition, TPN$_{gen}$ performs better than TPN$_{task}$ and a larger degree of improvement is attained when exploiting both general-purpose and task-specific adaptation by TPN. Please note that the highest accuracy 82.8\% of S-En is equipped with the test-time augmentation (Test-aug), i.e., averaged predictions of 16 different augmentations of each image, while the accuracy 80.4\% of our TPN is on single model without any DA. When relying on one kind of DA (Mini-aug), S-En only achieves 74.2\% which is still lower than ours.

\begin{figure*}[!tb]
    \centering {\includegraphics[width=1\textwidth]{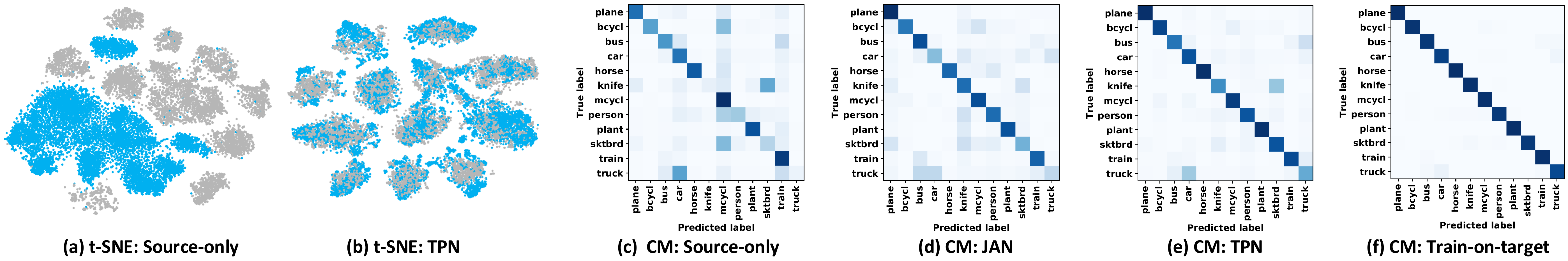}}
    %\vspace{-0.05in}
    \caption{\small (a)-(b): The t-SNE visualization of features generated by Source-only and TPN (gray: source, blue: target). (c)-(f): The Confusion Matrix (CM) visualization for Source-only, JAN, TPN and Train-on-target.}
    \label{fig:tsneconf}
    \vspace{-0.05in}
\end{figure*}

\begin{figure*}[!tb]
    \centering {\includegraphics[width=1\textwidth]{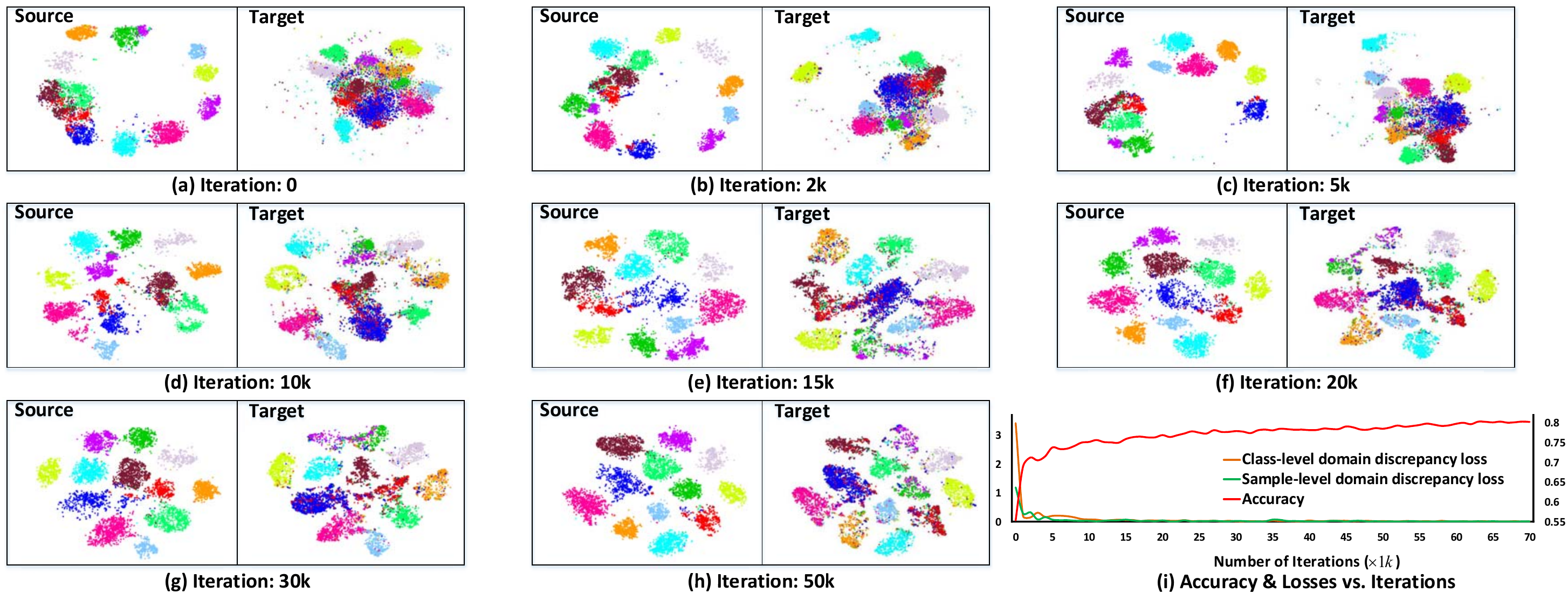}}
    %\vspace{-0.06in}
    \caption{\small (a)-(h): The t-SNE visualizations of features generated by TPN with the increase of the iteration on VisDA. (i): Accuracy \& Class-level and sample-level domain discrepancy losses with the increase of the iteration on VisDA (better viewed in color).}
    \label{fig:iter}
    \vspace{-0.1in}
\end{figure*}

\subsection{Experimental Analysis}\label{ssec:EA}
\textbf{Feature Visualization.}
Figure \ref{fig:tsneconf}(a)-(b) depict the t-SNE \cite{maaten2008visualizing} visualizations of features learnt by Source-only and our TPN on VisDA 2017 dataset (10$k$ samples in each domain). We can see that the distribution of target sample is far from that of source samples for Source-only run without domain adaptation. Through domain adaptation by TPN, the two distributions are brought closer, making the target distribution indistinguishable from the source one.

\textbf{Confusion Matrix Visualization.}
Figure \ref{fig:tsneconf}(c)-(f) show the visualizations of confusion matrix for the classifier learnt by Source-only, JAN, our TPN and Train-on-target on VisDA. Examining the confusion matrix of Source-only reveals that the domain shift is relatively large and the majority of the confusion are observed between objects with similar 3D structures, e.g., knife \& skateboard (sktbrd) and truck \& car. Through domain adaptation by JAN and TPN, the confusion is reduced for most classes. In particular, among all the 12 categories, TPN achieves higher accuracies than JAN for 10 categories, demonstrating that the features learnt by our TPN are more discriminative on target~domain.

\textbf{Convergence Analysis.}
To illustrate the convergence of our TPN, we visualize the evolution of the embedded representation of a subset on VisDA 2017 dataset (10$k$ samples for each domain) with t-SNE during training. Figure \ref{fig:iter}(a)-(h) illustrate that the target classes are becoming increasingly well discriminated by TPN source classifier. Figure \ref{fig:iter}(i) further depicts that the accuracy constantly increases (i.e., the noise of the pseudo labels $\rho$ decreases) and the two adaptation losses decrease when iterating more steps. Specifically, at the initial time, the ratio $\rho$ of target examples with false pseudo labels is 44.7\%, i.e., only 55.3\% of target samples are assigned with the correct labels. With the increase of training iterations of our TPN, such noise of pseudo labels $\rho$ is gradually decreased and the final accuracy will be boosted up to 80.4\% after model convergence. This again verifies that minimizing class-level and sample-level domain discrepancy will lead to better adaptation.

\section{Conclusions}
We have presented Transferrable Prototypical Networks (TPN), which explores domain adaptation in an unsupervised manner. Particularly, we study the problem from the viewpoint of both general-purpose and task-specific adaptation. To verify our claim, we have devised the measure of each adaptation in the framework of prototypical networks. The general-purpose adaptation is to push the prototype of each class computed in each domain to be close in the embedding space, resulting in invariant representation distribution across domains in general. The task-specific adaptation further takes the decisions of classifiers into account when aligning feature distributions, which ideally leads to domain-invariant representations. Experiments conducted on the transfers across MNIST, USPS and SVHN datasets validate our proposal and analysis. More remarkably, we achieve new state-of-the-art performance of single model on synthetic-to-real image transfer in VisDA 2017 challenge.

{\small
\bibliographystyle{ieee}
\bibliography{egbib}
}

\end{document}